\documentclass[letterpaper, 10 pt, conference]{ieeeconf}  % Comment this line out if you need a4paper

\IEEEoverridecommandlockouts                              % This command is only needed if 
\usepackage[english]{babel}
\usepackage{amsmath}
\usepackage{amssymb}
\usepackage{booktabs}
\usepackage{siunitx}
\usepackage[dvips]{graphicx}
\usepackage{multirow}
\usepackage{amsfonts}
\usepackage{enumerate}
\usepackage{tabularx}
\usepackage{algorithm,algorithmic}
\usepackage{bm}
\usepackage{enumerate}
\usepackage{adjustbox}
\usepackage{xcolor}
\usepackage{siunitx}
\usepackage{booktabs}
\usepackage{mdframed}
\usepackage{adjustbox}
\usepackage{authblk}
\usepackage{color}
\usepackage{xurl}
\usepackage{subcaption}
\usepackage{cite}
\makeatletter
\let\NAT@parse\undefined
\makeatother
\usepackage{hyperref}
\usepackage{relsize}
\usepackage{float}
\usepackage{pifont}

\title{\LARGE \bf
Trajectory Planning and Control for Differentially Flat\\ Fixed-Wing Aerial Systems 
}

\author{Luca Morando$^{1}$$^*$, Sanket A. Salunkhe$^{1}$$^*$, Nishanth Bobbili$^{1}$, Jeffrey Mao$^{1}$, Luca Masci$^{1}$, Hung Nguyen$^{2}$,\\ Cristino de Souza$^{2}$, and Giuseppe Loianno$^{1}$% <-this % stops a space

\thanks{$^*$Equal contribution and authors listed in alphabetical order.}
\thanks{$^1$The authors are with the New York University, Tandon School of Engineering, Brooklyn, NY 11201, USA. {\tt\footnotesize email: \{luca.morando, sas9908, nb3553, jm7752, lm5175, loiannog\}@nyu.edu}.}%
\thanks{$^2$The authors are with the Autonomous Robotics Research Center-Technology Innovation Institute, Abu Dhabi, UAE. {\tt\footnotesize email:  \{hung.tuan,cristino.dsouza\}@tii.ae}.}
\thanks{This work was supported by the Technology Innovation Institute, the NSF CAREER Award 2145277, and the DARPA YFA Grant D22AP00156-00, Qualcomm Research, Nokia, and NYU Wireless. Giuseppe Loianno serves as consultant for the Technology Innovation Institute. This arrangement has been reviewed and approved by the New York University in accordance with its policy on objectivity in research.}
}

\begin{document}

\maketitle
\thispagestyle{empty}
\pagestyle{empty}

%%%%%%%%%%%%%%%%%%%%%%%%%%%%%%%%%%%%%%%%%%%%%%%%%%%%%%%%%%%%%%%%%%%%%%%%%%%%%%%%
\begin{abstract}

Efficient real-time trajectory planning and control for fixed-wing unmanned aerial vehicles is challenging due to their non-holonomic nature, complex dynamics, and the additional uncertainties introduced by unknown aerodynamic effects. 
In this paper, we present a fast and efficient real-time trajectory planning and control approach for fixed-wing unmanned aerial vehicles, leveraging the differential flatness property of fixed-wing aircraft in coordinated flight conditions to generate dynamically feasible trajectories. The approach provides the ability to continuously replan trajectories, which we show is useful to dynamically account for the curvature constraint as the aircraft advances along its path. 
Extensive simulations and real-world experiments validate our approach, showcasing its effectiveness in generating trajectories even in challenging conditions for small FW such as wind disturbances.

% In this paper, we present an efficient real-time trajectory planning and generation framework for fixed-wing aircraft. This framework ensures dynamic feasibility, which is critical for non-holonomic systems like fixed-wings. Due to fixed-wings' continuous forward motion, we require a smooth transition between trajectories at waypoints while maintaining coupled dynamics and constant speed. In our approach, we represent our trajectory using a Bernstein polynomial with continuous replanning to respect non-linear coupled dynamics.  

\end{abstract}

%%%%%%%%%%%%%%%%%%%%%%%%%%%%%%%%%%%%%%%%%%%%%%%%%%%%%%%%%%%%%%%%%%%%%%%%%%%%%%%%

\section{Introduction}

In recent years, the deployment of small Fixed-Wing Unmanned Aerial Vehicles (FW-UAVs) has significantly increased across various applications, including environmental monitoring \cite{GREEN2019465}, low-altitude surveillance \cite{Jaimes}, and support for first responders in search and rescue operations \cite{Lyu}. Their popularity is primarily due to their superior endurance, extended operational range, and lower energy consumption compared to traditional Vertical Take-Off and Landing (VTOL) platforms like quadrotors.
Since FW-UAVs cannot hover in place or execute sharp turns and must maintain continuous motion to remain airborne, accurate trajectory planning and precise tracking are essential for their safe operations.  
Given the intrinsic nonlinear and coupled translational and rotational dynamics of FW-UAVs, shown by their non-holonomic nature and the requirements of applying a rolling motion to change their heading,  the generated trajectory must exhibit 
$\mathcal{C}^3$ continuity and be planned to keep the system's states within a safe operating envelope \cite{Tekles, Johannes}.  \begin{figure}[t!]
\includegraphics[width=\columnwidth,  trim=0cm 5cm 0cm 0cm, clip]{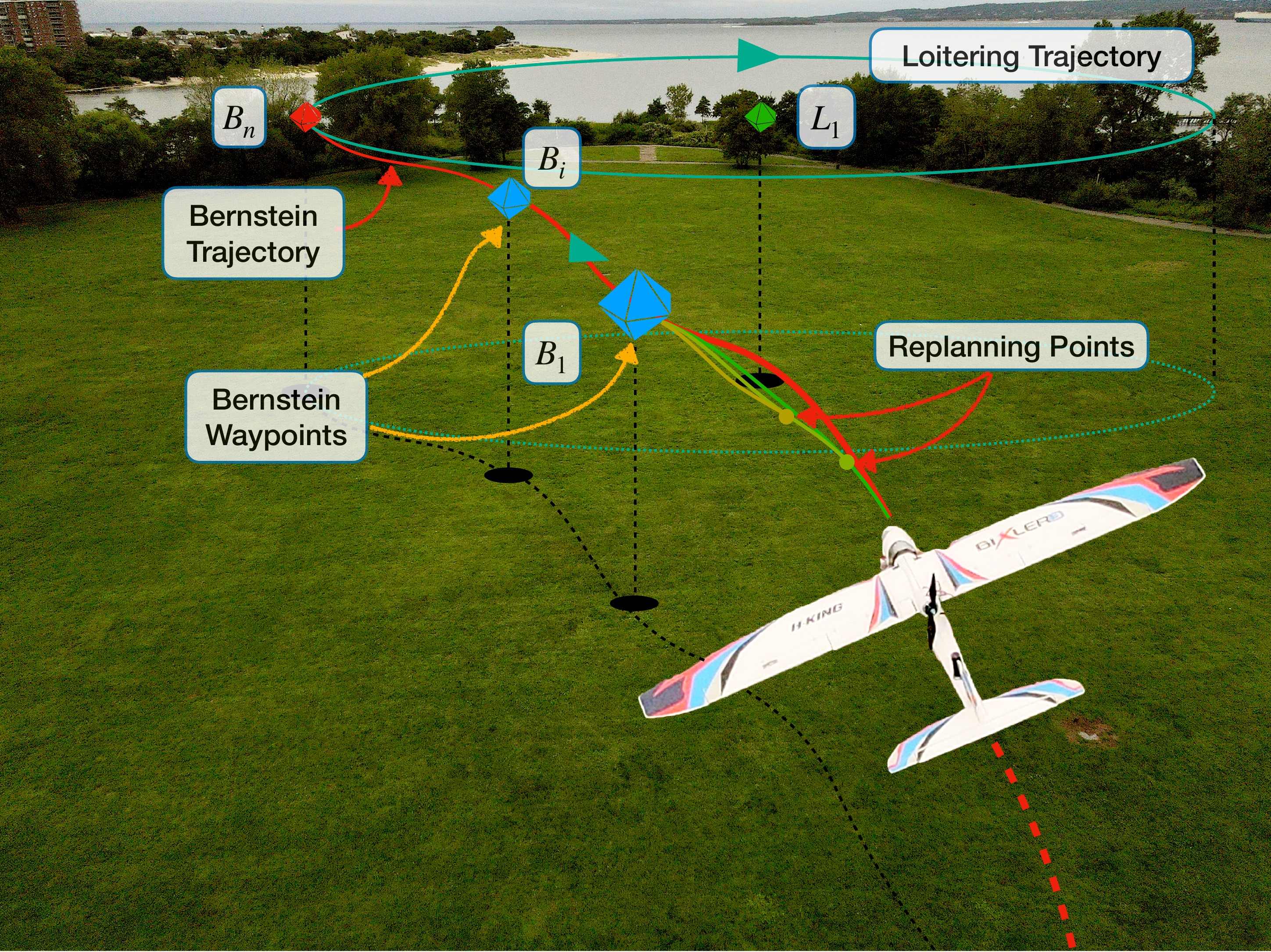}
  \caption{Continuous, online trajectory replanning between multiple waypoints during a sample real-world flight.
}
  \label{fig:fig1}
\vspace{-20pt}
\end{figure}

Several solutions often bypass the complexity of flight dynamics and instead focus solely on kinematic models relaxing the $C^3$ continuity or compute short trajectories directly in the system's states, resulting in a high computational burden \cite{Barry, Levin}. In case the coordinated flight condition is respected \cite{hauser1997aggressive}, FW-UAVs can be identified as a differentially flat system similar to multicopters \cite{Mellinger}.  Leveraging the system's differential flatness property, the planning problem can be simplified by directly mapping the flat output variables and their higher derivatives in the system's state space, obtaining the desired input values required to follow the trajectory  \cite{Bry2015AggressiveFO}.

In this work, we propose a novel strategy for real time dynamically feasible trajectory planning and control for FW-UAV. 
Differently from previous works, which rely on offline, computationally inefficient suboptimal optimization~\cite{Guozheng} or optimization on top of primitive Dubins-polynomial curves \cite{LugoCrdenas2014DubinsPG}, our approach is entirely based on Bernstein polynomials~\cite{kielas2022bernstein}, determining the corresponding coefficients via convex quadratic optimization, offering a more efficient and scalable method for real-time trajectory planning. The contribution of the paper can be summarized as
\begin{itemize}
\item We propose a novel trajectory planning and control approach that leverages the differential flatness of FW-UAVs. Our method effectively bridges the gap between theory and practice, providing a clear formulation that can be implemented on real robots.

\item We demonstrate continuous trajectory replanning, which we show is helpful to dynamically adjust the curvature constraint as the UAV advances along its path.

\item We present simulations and real-world results that showcase the effectiveness of our approach, even in challenging conditions for small FW such as wind disturbances, enabling efficient real-time onboard computation of trajectories spanning hundreds of meters.

\end{itemize}
\section{Related Works}
\label{Sec:Related_Works}
\vspace{-8pt}
Compared to small purely VTOL rotorcraft like quadrotors, FW-UAVs provide several advantages, such as longer flight endurance, lower energy consumption, and the capacity to carry heavier payloads. However, these advantages come at the cost of increased state-coupled dynamics complexities affected as well by unknown aerodynamics effects~\cite{Dobrokhodov2020}. 
Therefore, quickly generating and tracking dynamically feasible trajectories present a significant challenge. Most existing trajectory generation algorithms for FW-UAVs simplify the problem by focusing on kinematic models, bypassing the intricate flight dynamics. In \cite{Chitsaz} an extension of the Dubins path is used to compute a time-optimal trajectory with curvature constraints. A similar approach is found in \cite{ryu2023path}, where Dubins-based motion primitives are modified to incorporate smoother transitions between segments. In \cite{Wang},  the trajectory generation is treated as a kinematic planning problem, connecting lines and arcs of constant curvature. However, when tested in simulation, this method leads to instantaneous accelerations at segment connections, causing dynamic instabilities during trajectory tracking.
Similarly, in \cite{Frazzoli}, a set of dynamically feasible "trim primitives" are concatenated to create a complex motion plan. However, these methods, primarily based on Dubins paths, fail to provide $\mathcal{C}^3$ continuity at the segment junctions \cite{Gros, Johannes}.
Other methods compute short trajectories directly in the system’s states~\cite{Barry, Levin}, imposing high computational costs due to nonlinear flight dynamics. In trajectory generation, differential flatness~\cite{Nieuwstadt} enables transformation from flat output space to state and control input space~\cite{Martin}, widely applied in quadrotors~\cite{Mellinger} for aerobatic maneuvers.

For fixed-wing aircraft, \cite{hauser1997aggressive} introduces differential flatness under coordinated flight, later utilized in~\cite{Bry2015AggressiveFO, Liu}. However, unlike \cite{Liu, Guozheng}, which use offline trajectory generation with MINCO~\cite{wang2022geometrically} and test only in simulation, we propose a fast, online method optimizing Bernstein polynomials~\cite{kielas2019bebot} up to the third derivative. Unlike SP-line~\cite{Johannes_book, Tal} or nonlinear Bezier-based methods~\cite{Celestini}, our approach exploits Bernstein polynomial properties~\cite{kielas2019bebot} for efficient online trajectory generation via quadratic optimization~\cite{GertzOSQP}, ensuring smooth, continuous tracking.

As in~\cite{hauser1997aggressive, Bry2015AggressiveFO}, our trajectories are parameterized in space to maintain constant cruising velocity. Unlike~\cite{Bry2015AggressiveFO, Liu}, we validate in challenging outdoor conditions with small FW aircraft and deploy in real-time. This work bridges theory and practice, providing a clear formulation that can be implemented on real robots.

\begin{figure}[t!]
\includegraphics[width=0.8\columnwidth, trim={1cm 1cm 3cm 0.5cm}, keepaspectratio]{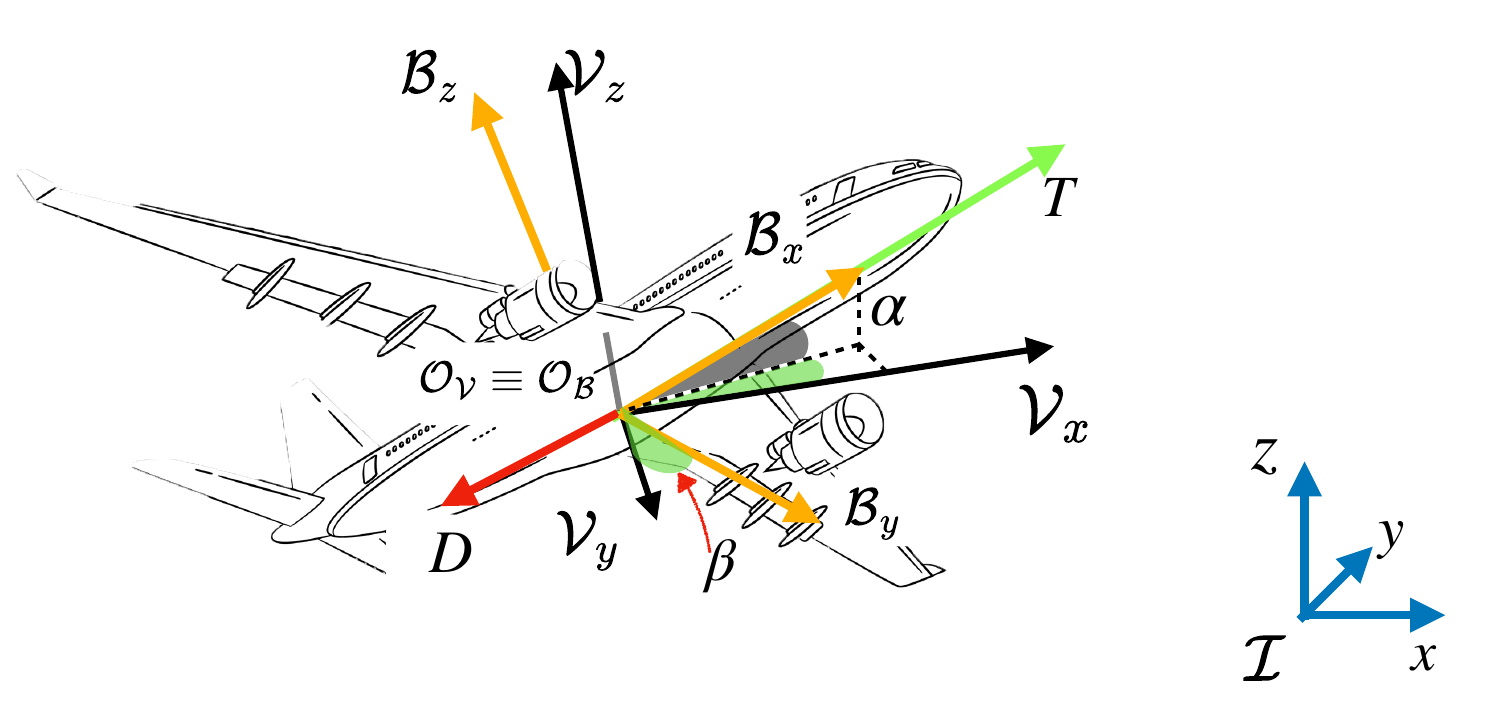}
  \caption{Frames' visualization and convention. The Velocity frame $\mathcal{V}$ differs from the Body frame $\mathcal{B}$ by the angles $\alpha$ and $\beta$, whereas $\mathcal{I}$ defines the Inertial fixed frame. 
}
 \label{fig:fig2}
\vspace{-10pt}
\end{figure}

\begin{figure*}[!t]
  \centering
  \includegraphics[width=1\textwidth]{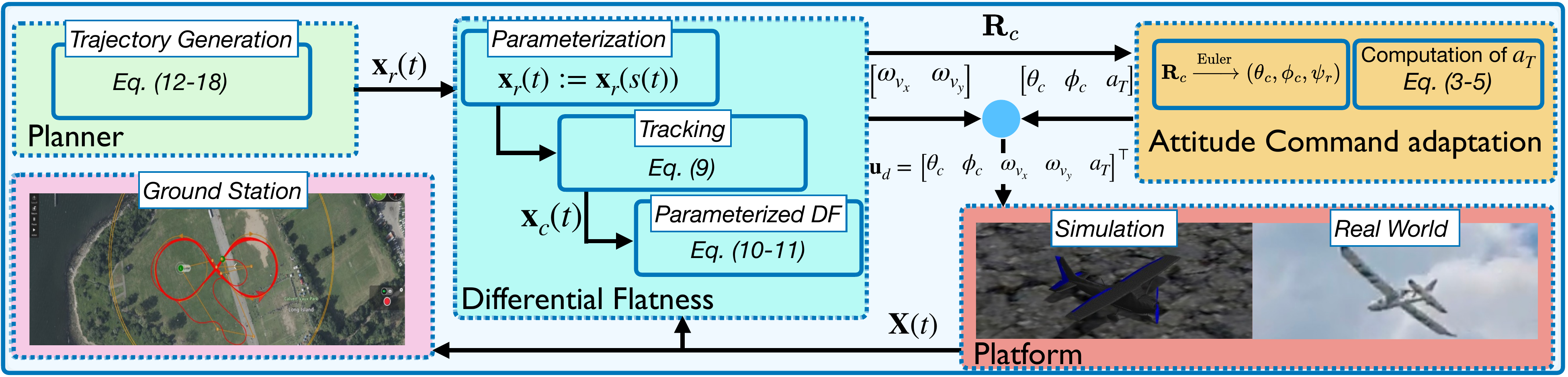}
  \caption{Architecture overview. The trajectory generated by our planner is forwarded to the Differential Flatness Block, which computes the desired inputs to control the attitude and the longitudinal thrust of the simulated or real robot. %The mission is monitored through the Ground Station. 
  \label{fig:software_architecture}}
  \vspace{-20pt}
\end{figure*}

\section{System Modeling}
\label{sec:System_modeling}

In this section, we outline a simple aircraft model that captures the important features of the coordinated flight condition adopted in the rest of this paper.
As shown in Fig.~\ref{fig:fig2}, we use the following frame convention: the inertial frame is denoted by $\mathcal{I}$ while $\mathcal{B}$ denotes the vehicle’s rigid body frame that it is aligned with the FW's longitudinal, lateral, and vertical axes. The velocity frame $\mathcal{V}$, centered at $\mathcal{B}$, is rotated with respect to the $\mathcal{B}$ by the sideslip angle $\beta$ and the attack angle $\alpha$, therefore always keeping the corresponding $\mathcal{V}_x$ axis aligned with the direction of the aircraft velocity. The purpose of introducing the velocity frame $\mathcal{V}$ is due to the fact that the FW can be subject to lateral winds that may deviate its nose from the desired direction of motion. 
The state of the FW system is defined as $\mathbf{X} = \{\mathbf{x}, \mathbf{\dot{x}}, \mathbf{R}, \bm{\omega}_v\}$, where $\mathbf{x} \in \mathbb{R}^3$ represents the position of FW in inertial frame $\mathcal{I}$,  
%The rotation between $\mathcal{B}$ and  $\mathcal{V}$, is represented by $R_{\mathcal{V}}^{\mathcal{B}}(\alpha, \beta)$, 
while $\mathbf{R} \in SO(3)$ with $\mathbf{R}= \mathbf{R}_{\mathcal{B}}^{\mathcal{I}}\mathbf{R}_{\mathcal{V}}^{\mathcal{B}}$ denotes the rotation of $\mathcal{V}$ with respect to $\mathcal{I}$. This can also be parameterized using Euler angles roll ($\phi$), pitch ($\theta$), and yaw ($\psi$). The velocity of fixed-wing is represented by the velocity vector $\mathbf{\dot{x}} = [\dot{x}_{x}~\dot{x}_{y}~\dot{x}_{z}]^\top$, with a zero lateral velocity component ($\dot{x}_{y} = 0$) to satisfy the coordinated flight condition. This condition is general to different aircraft configurations, assuming that the motion of the plane is aligned with the relative wind direction and executes curves keeping the lift vector always aligned with $\mathcal{V}$ vertical axis. 

%The orientation dynamics can be described by equation $\dot{R} = R\hat{\mathbf{\omega}}_{b}$, where $\hat{\mathbf{\omega}}_{b}$ is the skew-symmetric matrix of the instantaneous angular velocity vector $\boldsymbol{\omega}_b = [\omega_{b_x}, \omega_{b_y}, \omega_{b_z}]^T $ expressed in the body frame. 

The system dynamics can be described as
\begin{equation}
    \begin{split}
    \mathbf{\dot{x}} = V\mathbf{R}\mathbf{e}_1,&~\mathbf{\ddot{x}} = \mathbf{g} + \mathbf{R}\mathbf{a}_v,\\
    \mathbf{\dot{R}} = \mathbf{R}\hat{\bm\omega}_v,~\dot{\bm{\omega}}_v &= \mathbf{J}^{-1}(-\bm{\omega}_v \times \mathbf{J}\bm{\omega}_v + \bm{\tau}),\label{eq:system_dynamics4}
\end{split}
\end{equation}
where $V = \lVert \dot{\mathbf{x}} \rVert$, $\mathbf{g} = [0~0~-g]^{\top}$ is the gravity vector in $\mathcal{I}$, $\mathbf{e}_1 = [1~0~0]^{\top}$ is the versor aligned with the $x$ direction of $\mathcal{V}$, $\mathbf{a_v} = [a_{v_x}~0~a_{v_z}]^\top$ contains the axial and normal accelerations, while $\boldsymbol{\omega}_v = [\omega_{v_x}~\omega_{v_y}~ \omega_{v_z}]^\top$ represents the angular velocity of $\mathcal{V}$ wrt. $\mathcal{I}$, expressed in $\mathcal{V}$ and $\hat{\bm\omega}_v$ its corresponding skew-symmetric matrix. Finally, $\mathbf{J} \in \mathbb{R}^{3 \times 3}$ describes the inertia acting on each direction of the body frame $\mathcal{B}$, while $\bm{\tau}$ expresses the torque applied on the system due to the action of control surfaces, like ailerons, elevators, and rudder. 
Moreover, as described in \cite{hauser1997aggressive}, to maintain coordinated flight conditions, the second and third components of the angular velocity $\boldsymbol{\omega}_v$, are constrained to be 
\begin{equation}
    \omega_{v_y} = -(a_{v_z} + g_{v_z})/V,~\omega_{v_z} = g_{v_y}/V,\label{eq:system_dynamics5}
\end{equation}
where $\mathbf{g}_{v} = \mathbf{R}^{\top} \mathbf{g}$.
Therefore, the coordinated flight conditions do not impose any constraints on $\omega_{v_x}$ of the FW. 

\subsection{Aerodynamics and Propulsion Model}
\label{sec:Aerodynamics_and_prop}
%The coordinated flight model is a so powerful mathematical notation, that despite being very concise, can describe the translational dynamics and attitude kinematics of a fixed wing aircraft, leveraging normal and axial accelerations, respectively $a_{v_x}$ and $a_{v_z}$ and the roll velocity $\omega_1$. 

More realistic attitude dynamics can be obtained by also including the acceleration due to lift, drag, and thrust, respectively $a_L$, $a_D$, and $a_T$, which are generally modeled as a function of the altitude $x_z$, airspeed $V_a$, and angle of attack $\alpha$. In this paper, we consider $a_L$, $a_D$, and $a_T$  to be~\cite{tseng1988calculation}

\begin{align}
a_L &= \frac{\sigma(x_z)V_a^2SC_L}{2m} + a_{L,0}, \\
a_D = &\frac{\sigma(x_z) V_a^2 S C_{d}}{2m},~a_T = T/m + a_D,
\end{align}
where $\sigma$, $S$, $T$, and $m$ are the air density, wing surface area, motor thrust and mass of FW. The lift coefficient $C_L$, initial lift acceleration $a_{L,0}$, and drag coefficient $C_D$ depend on the aerodynamic properties of the aircraft, including its shape and angle of attack. 
Therefore, the axial and normal acceleration inputs to the system are given by
\begin{equation}
    a_{v_x} = a_T \cos{\alpha} - a_D,~a_{v_z} = -a_T \sin{\alpha} - a_L.
\end{equation}

\subsection{Differential Flatness}
\label{sec:differential_flat}
%Adding here a general introduction to diff flat and how to find R des given the versors
%Mathematically, this can be expressed as:
%\begin{align}
%\dot{\mathbf{X}} &= f(\mathbf{X}, \mathbf{u}) \\
%(\mathbf{X}, \mathbf{u}) &= \Psi(z, \dot{z}, ..., z^i) 
%\end{align}
%where $\Psi(\dots)$ denotes the mapping function between the flat outputs $z$ and the system states $\mathbf{X}$ and the inputs $\mathbf{u}$, with $\text{dim}(\mathbf{z}) = \text{dim}(\mathbf{u})$.

This section provides an overview of the Differential Flatness and Feedback Trajectory Tracking blocks shown in Fig. \ref{fig:software_architecture}. A system is considered differentially flat if there exists a set of of flat outputs, such that the system's state and input can be fully described in terms of these outputs and their derivatives.
In the case of an FW system operating under the coordinated flight equations introduced in Section~\ref{sec:System_modeling}, it becomes a feedback linearizable system~\cite{hauser1997aggressive}, where the flat output is represented by the position $\mathbf{x}$, while the inputs to the model are $\mathbf{u} = [\dot{a}_{v_x}~\omega_{v_x}~\dot{a}_{v_z}]^\top$. 
Following \cite{hauser1997aggressive} and differentiating the acceleration expression $\ddot{\mathbf{x}}$ in eq. \eqref{eq:system_dynamics4}, we obtain $\mathbf{x}^{(3)} = \mathbf{R}(\bm{\omega}_v  \times \mathbf{a}_v + \dot{\mathbf{a}}_v)$, which is equivalent to
\begin{align}
\mathbf{x}^{(3)} &= 
\begin{bmatrix}
\omega_{v_y} a_{v_z} \\
\omega_{v_z} a_{v_x} \\
-\omega_{v_y} a_{v_x}
\end{bmatrix}
+ \mathbf{R}
\begin{bmatrix}
\dot{a}_{v_x} \\
-\dot{a}_{v_z}\omega_{v_x} \\
-\dot{a}_{v_z} 
\end{bmatrix}.
\end{align}

Inverting the following expression directly lead to the final differential flatness equation

%Diff flat equation 71 from IJRR paper 
\begin{equation}
\begin{aligned}
\begin{bmatrix}
\dot{a}_{v_x} \\
\omega_{v_x} \\
\dot{a}_{v_z}
\end{bmatrix} = 
\begin{bmatrix}
-\omega_{v_y}a_{v_z} \\
\omega_{v_z}a_{v_x}/a_{v_z} \\
\omega_{v_y}a_{v_x}
\end{bmatrix} + 
\begin{bmatrix}
1 & 0 & 0 \\
0 & -1/a_{v_z} & 0 \\
0 & 0 & 1
\end{bmatrix} \mathbf{R}^\top
\mathbf{x}^{(3)},
\end{aligned}
\label{eq:differential_flatness}
\end{equation}
where $\mathbf{R}=[\mathbf{r}_{x}~\mathbf{r}_{y}~\mathbf{r}_{z}]$ with $\mathbf{r}_{x} = \dot{\mathbf{x}}/\lVert{\dot{\mathbf{x}}}\rVert$, 
$\mathbf{r}_{z} = \mathbf{a}_{n} / a_{v_z}$, and $\mathbf{r}_{y} = \mathbf{r}_{z} \times \mathbf{r}_{x}$. 
Therefore, $a_{v_z} = - \lVert{\mathbf{a}_{n}}\rVert $ where $\mathbf{a}_{n}$ is found by the projection of $\ddot{\mathbf{x}}$ in the normal plane as $\mathbf{a}_{n} = (\ddot{\mathbf{x}}- \mathbf{g} - a_{v_x} \mathbf{r}_{v_x})$, where $a_{v_x} = \mathbf{r}_{x}^\top (\ddot{\mathbf{x}} - \mathbf{g})$. To respect the coordinated flight condition $a_{v_y} = 0$. The differential flatness equation only holds if the flatness constraints, namely $\mathbf{\dot{x}} \neq 0$ and $a_{v_z} \neq 0$, are satisfied. This is intuitive, as the aircraft's lack of hovering capability and the inability to control the system when the aircraft is perpendicular to the desired trajectory direction make these constraints necessary.
%makes possible to find the axial acceleration $a_{v_x} = \mathbf{r}_{v_x}^T (\ddot{\mathbf{x}} - \mathbf{g}) $. The normal acceleration $a_{v_z}$ is found by the projection of $a_{v_x}$ in the normal plane as  $a_{v_z} = -\lVert(\ddot{\mathbf{x}\rVert - g - a_{v_1} \mathbf{r}_{v_x}^T)}$. Thus the full matrix $R$, describing the desired attitude of $\mathcal{V}$ given the input $\mathbf{z}^{(3)}$ can be retrieve knowing that $\mathbf{r}_{v_z} = a_{v_z} / \lVert a_{v_z}\rVert$ and $\mathbf{r}_{v_y} = \mathbf{r}_{v_z} \times \mathbf{r}_{v_x}$, since no accelerations are wanted on $\mathcal{V}_y$.

We define the system's control input sent to the inner attitude controller~\cite{REINHARDT202191,Coates} the desired orientation matrix $\mathbf{R}_{c}$, expressed through Euler angles $\theta_c$, $\phi_c$, and $\psi_c$, along with angular velocities $\omega_{v_x}$, $\omega_{v_y}$, and axial acceleration $a_{v_x}$ represented as thrust $a_T$. This forms the commanded control input $\mathbf{u}_{c} = [\theta_c~\phi_c~\omega_{v_x}~\omega_{v_y}~a_T]^\top$. Specifically, $\mathbf{u}_{c}$ is derived by first calculating $\mathbf{R}_c$ from the previous $\mathbf{R}$ expression. Subsequently, we consider the following cascade PID loop to compute the commanded jerk
\begin{equation}
    \mathbf{x}_{c}^{(3)} = \mathbf{x}^{(3)}_{r} + k_2 \ddot{\mathbf{e}} + k_1 \dot{\mathbf{e}} + k_0 \mathbf{e},
\end{equation}
where $\mathbf{e} = \mathbf{x}_{r}(t) - \mathbf{x}(t)$, and $k_2, k_1, k_0$ the feedback gains. Finally, based on the differential flatness model in eq. \eqref{eq:differential_flatness},  we derive $\omega_{v_x}~\text{and~}\omega_{v_y}$ considering $\mathbf{x}_{c}^{(3)}$ and $\mathbf{R}_c$ in place of $\mathbf{x}^{(3)}$ and $\mathbf{R}$ respectively. This allows to achieve a trajectory tracking given the state feedback $\mathbf{x}(t)$.

\subsection{Trajectory Time Parametrization}
Despite the strength of the differential flatness approach, the  desired tangential acceleration along the trajectory can vary depending on how the trajectory is formulated with respect to time. Due to the natural minimization of the tracking error $\mathbf{e}$ towards the reference trajectory tracking point, an abrupt change of the desired thrust may happen if the trajectory presents variations in the reference velocities $\dot{\mathbf{x}}_{r}$ and acceleration $\ddot{\mathbf{x}}_{r}$.
In this condition, the FW can slow down below a safe cruising airspeed, producing a loss of airflow and control of the aerodynamic surfaces. 

To prevent such a scenario, we introduce a path parameterization variable $s(t)$ that defines how the desired trajectory values are allocated along the path $\mathbf{x}_{r}:= \mathbf{x}_{r}(s(t))$ for $t \geq 0$, where $s$ represents the distance along the desired path. This parameterization enables dynamic inversion of trajectory $\mathbf{x}_{r}(t) $ based on the distance travelled while maintaining a constant cruising velocity. Therefore, eq.~\eqref{eq:differential_flatness} is modified as

%ensures that the axial acceleration $a_{v_x}$ remains constant and predetermined for the entirety of the trajectory; however, it also means that $a_{v_x}$ is no longer available as control input in equation \ref{eq:differential_flatness}. Thus, apath parameterization $s(t)$ variable is introduced, which follows the desired trajectory $\rho_s(t):= \rho(s(t))$ for $t \geq 0$, where $s$ is the distance along the desired path. This path parameterization enables dynamic inversion of trajectory $\rho(t)$ based on the distance travelled while maintaining a constant cruising velocity. It also provides an alternative state representation where $\dot{a}_{v_x}$ can be expressed as $s^{(3)}$, allowing us to modify the differential flatness equation \ref{eq:differential_flatness} in \cite{hauser1997aggressive} as shown below:

\begin{equation}
\begin{split}
& \mathbf{M}
\begin{bmatrix}
s^{(3)} \\
\omega_{v_x} \\
\dot{a}_{v_z}
\end{bmatrix} = 
\begin{bmatrix}
a_{v_z}\omega_{v_y} + \dot{a}_{v_z}\\
a_{v_x}\omega_{v_z} \\
-a_{v_x}\omega_{v_y}
\end{bmatrix} \\
 &- 
\mathbf{R}^\top \left[3\frac{\delta^2\mathbf{x}_{r}}{\delta s^2}\ddot{s}\dot{s} + \frac{\delta^3\mathbf{x}_{r}}{\delta s^3}\dot{s}^3 + k_2\mathbf{\dddot{e}} + k_1\mathbf{\dot{e}} + k_0\mathbf{e}\right],
\end{split}
\label{eq:time_param_differential_flatness}
\end{equation}
where $\mathbf{M}$ is the decoupling matrix represented as
\begin{equation}
\mathbf{M}
 = 
\begin{bmatrix}
\vdots & 0 & 0\\
\mathbf{R}^\top\frac{\delta\mathbf{x}_{r}}{\delta s}& a_{v_z} & 0 \\
\vdots & 0 & -1
\end{bmatrix}. 
\label{eq:time_param_M}
\end{equation} 

\section{Trajectory Planning}
\label{sec:Planning}

% problem definition for motion planning 
%In the previous section, we introduced the differentially flat model for FW constrained to a coordinated flying condition. 
We focus on designing an optimal, dynamically feasible trajectory for a FW that leverages the differential flatness property and employs Bernstein polynomials adhering to the following conditions

 %Using this concise yet powerful representation, it is possible to obtain the desired input $\mathbf{u}$ to the vehicle directly from the third derivative of the desired trajectory $\rho(t)$, which has to be continuous and differentiable, in order to guarantee smooth roll rate control. 
 
\begin{itemize}
    \item The axial velocity of the plane $\dot{x}_{x} \neq 0$.
    \item The trajectory should satisfy that $a_{v_z} \neq 0 $.
    \item Bounding the maximum curvature $\kappa$ of the trajectory.
\end{itemize}
We formulate a convex quadratic optimization of Bernstein polynomials to minimize the trajectory jerk $\mathbf{x}_{r}^{(3)}(t)$ (input in eq. \eqref{eq:differential_flatness}) while keeping velocity, acceleration and curvature $\kappa$ constraints within specified bounds. %Since the FW aircraft are constrained to continuous forward motion.%, each Bernstein trajectory is followed by a circular loitering motion with a predetermined radius when no new trajectory is defined. This loitering phase serves as the fixed-wing equivalent of the hovering phase seen in quadrotors or tail-sitters, as described in \cite{lu2024trajectory}.

%that will span  entire flight is significantly challenging compared to multi-rotor platforms \cite{loianno2016estimation}. Unlike quadrotors or tail-sitters \cite{lu2024trajectory}, which can hover in place, fixed-wing must continue its forward motion. This constraint forces the fixed wing either to immediately transition into a loiter motion or begin a new trajectory as soon as it completes its current one. In addition to a smooth transition between trajectories, it is important to respect the kino-dynamic constraint of a fixed-wing while generating these trajectories, particularly its minimum turning radius, which directly influences the required roll angle during flight. Respecting these kino-dynamic constraints and ensuring smooth transitions between trajectories is important to maintain stability and control throughout the entire flight. 

% solution discussion about a seamless transition
%To ensure the seamless transition between trajectories, we generate continuous and dynamically feasible trajectories within the vehicle's flat output space, defined by the vector $\mathbf{x}$. Throughout the entire flight span, the vehicle's trajectory is composed of three distinct types of polynomial trajectories: i) line trajectory, ii) loiter trajectory, and iii) Bernstein Trajectory. Each of these trajectories is defined by a sequence of flat outputs, which correspond to the desired positions and their higher derivatives up to the jerk at each time $t$ from the initial time $t_0$ to a final time $t_f$ for the respective trajectory.

\subsection{Bernstein Trajectory}
\label{sec:planner}

A Bernstein polynomial shows interesting properties in terms of smoothness and ability to impose global spatial constraints compared to time-based polynomials~\cite{kielas2019bebot,kielas2022bernstein}. For a given $m_j$  trajectory, it can be described by the following form of degree $n$
\begin{equation}
C_{n,m_j}(t) = \sum_{i=0}^{n}\mathbf{p}_{i,n}^{m_j}\beta^n_i(t),   \quad t\in [t_0, t_f]
\label{eq:bernstein_equation}
\end{equation}
where $\mathbf{p}_{i,n}^{m_j}$ are the Bernstein coefficient or control points of  size $n$ control, and $\beta^n_i(t)$ is the Bernstein basis. The $k^{\text{th}}$ derivative of the polynomial can be obtained as
\begin{equation}
\frac{d^{k}}{dt^{k}}C_{n,m_j}(t) = \frac{n!}{(n-k)!(t_f - t_0)^k} \sum_{i=0}^{n-k} {{\mathbf{p}}^{{m_j}^{'}}_{i,n-k}}\beta_{i}^{n-k}(t),
\label{eq:bernstein_derivative}
\end{equation}
with ${\mathbf{p}}^{{m_j}^{'}}_{i,n-k} = \mathbf{p}_{i,n}^{m_j}\mathbf{D}_k$ and $\mathbf{D}_k = \text{diag}(\mathbf{c}\circledast^k, \mathbf{c}\circledast^k, \cdots, \mathbf{c}\circledast^k)$ is the Differential matrix with $ \mathbf{c} = [-1, 1]$ convoluted $k$ times. Considering $M+1$ waypoints, a full trajectory $\mathbf{x}_{r}(t)$ can be modeled by stacking together $M$ Bernstein polynomials connected at the extremal points as

%A trajectory $\rho(t)$ can be efficiently modeled as a piece-wise stacking of consecutive Bernstein polynomial segments passing through multiple waypoints, combined into a single optimized Bernstein polynomial and connected at the extremes in order to be continuous-time in the flat output space $\mathbf{x}=\{x, y, z\}$ and its higher derivatives. 

%These trajectories can be efficiently optimized and evaluated under various constraints, including velocity, acceleration, and, especially in our case, roll rate and turning radius. By representing continuous-time trajectories as piecewise Bézier curves passing through multiple waypoints, these can be combined into a single Bernstein polynomial after optimization, ensuring both computational efficiency and respect the imposed constraints.

 %The general equation of Bernstein polynomial trajectory for a single dimension is represented as \cite{kielas2022bernstein}
%\begin{equation}
%C_n(t) = \sum_{i=0}^{n}\mathbf{p}_{i,n}\beta^n_i(t),   \quad t\in [t_0, t_f]
%\label{eq:bernstein_equation}
%\end{equation}

%where $\mathbf{p}_{i,n}$ are the Bernstein coefficient or control points, with each segment having a total of $n$ control points, and $\beta^n_i(t)$ is the Bernstein basis. To represent the complete trajectory passing through $M+1$ waypoints, we employ a set of $M$ Bernstein polynomials, where $M$ is the total number of trajectory segments:

\begin{equation}
\mathbf{x}_{r}(t)  = 
\begin{cases} 
    \sum_{i=0}^{n}\mathbf{p}_{i,n}^{m_1}\beta_i^n(T_1 - t) \quad \text{for} \ t\in [0, T_1]\\
    \sum_{i=0}^{n}\mathbf{p}_{i,n}^{m_2}\beta_i^n(T_2 - t) \quad \text{for} \ t\in [T_1, T_2] \\
    \vdots \\ 
    \sum_{i=0}^{n}\mathbf{p}_{i,n}^{M}\beta_i^n(T_{M}-t) \quad \text{for} \ t\in [T_{M-1}, T_M]
\end{cases}
\label{eq:piecewise_bernstein_equation}
\end{equation}
where $\mathbf{p}_{i,n}^{m_j}$ is the $i^{th}$ control point of the $m_j$ sub trajectory, with $j \in [1, M]$, and the time instants $T_1, T_2, \dots, T_M$ represent the allocated time for each of sub trajectory. 

%\begin{itemize}
%\item $P^{'}_{i,n-m} = P_{i,n}\mathbf{D}_m$
%\item Differential Matrix: $\mathbf{D}_m = \begin{bmatrix}\mathbf{s}\circledast^m  & 0 & \cdots  & 0 \\ 0 & \mathbf{s}\circledast^m & \cdots  & 0 \\\vdots  & \vdots  & \ddots  & \vdots  \\0 & 0 & \cdots  & \mathbf{s}\circledast^m\end{bmatrix}$
%\item $\mathbf{s} = [-1, 1] \quad \text{and} \quad \underbrace{s \circledast s \circledast \cdots \circledast s}_{m}=\mathbf{s}\circledast^m$
%\end{itemize}

To find the Bernstein Coefficients $\mathbf{p}$ we formalize  a Convex Quadratic Programming (QP) problem \cite{mao2023robust}
\begin{equation}
\begin{aligned}
\text{min} \quad & \mathbf{p}_d^T\mathbf{Q} \mathbf{p}_d\\
\text{s.t.} \quad & \mathbf{A}_{eq}\mathbf{p}_d = \mathbf{b}_{eq} \\
& \mathbf{A}_{ineq}\mathbf{p}_d \le  \mathbf{b}_{ineq}
\end{aligned}
\end{equation}
where $\mathbf{Q} = \text{diag}(Q_1, \hdots, Q_M)$ with $Q_i \in \mathbb{R}^{n \times n}$ representing the Hessian semi-definite matrix of the objective function, related to the $n$ number of Bernstein Coefficients each sub trajectory. The vector $\mathbf{p}_d$, with dimension $M \times n$, contains the Bernstein coefficients to be optimized for each spatial dimension $d$. To ensure continuity in position and higher derivatives between the segments, the optimization problem is subject to various equality and inequality constraints, which are represented by the matrices $\mathbf{A}_{eq}, \mathbf{A}_{ineq}$, and vectors $\mathbf{b}_{eq}, \mathbf{b}_{ineq}$

% Matrix $\mathbf{A}_{eq}, \mathbf{A}_{ineq}$, and vector $\mathbf{b}_{eq}, \mathbf{b}_{ineq}$ are derived from the equality and inequality constraints imposed by user for each dimension $d$. 
%The optimization problem is solved using an off-the-shelf OOQP \cite{gertz2003object} convex solver. 

% The vectors $\mathbf{b}_{id}$ consists of all number $i$ of constraints imposed by the user for each dimension $d$. Finally the matrix $\mathbf{A}  = \text{diag}(A_1, \hdots, A_j, \hdots, A_m)$ is composed by submatrix $A_j$ which one with dimension $A_j \in \mathbb{R}^{i \times n}$ and it is stacked for the $d$ dimesions of the polynomial.

\begin{figure}[!t]
  \centering
  \includegraphics[width=\columnwidth]{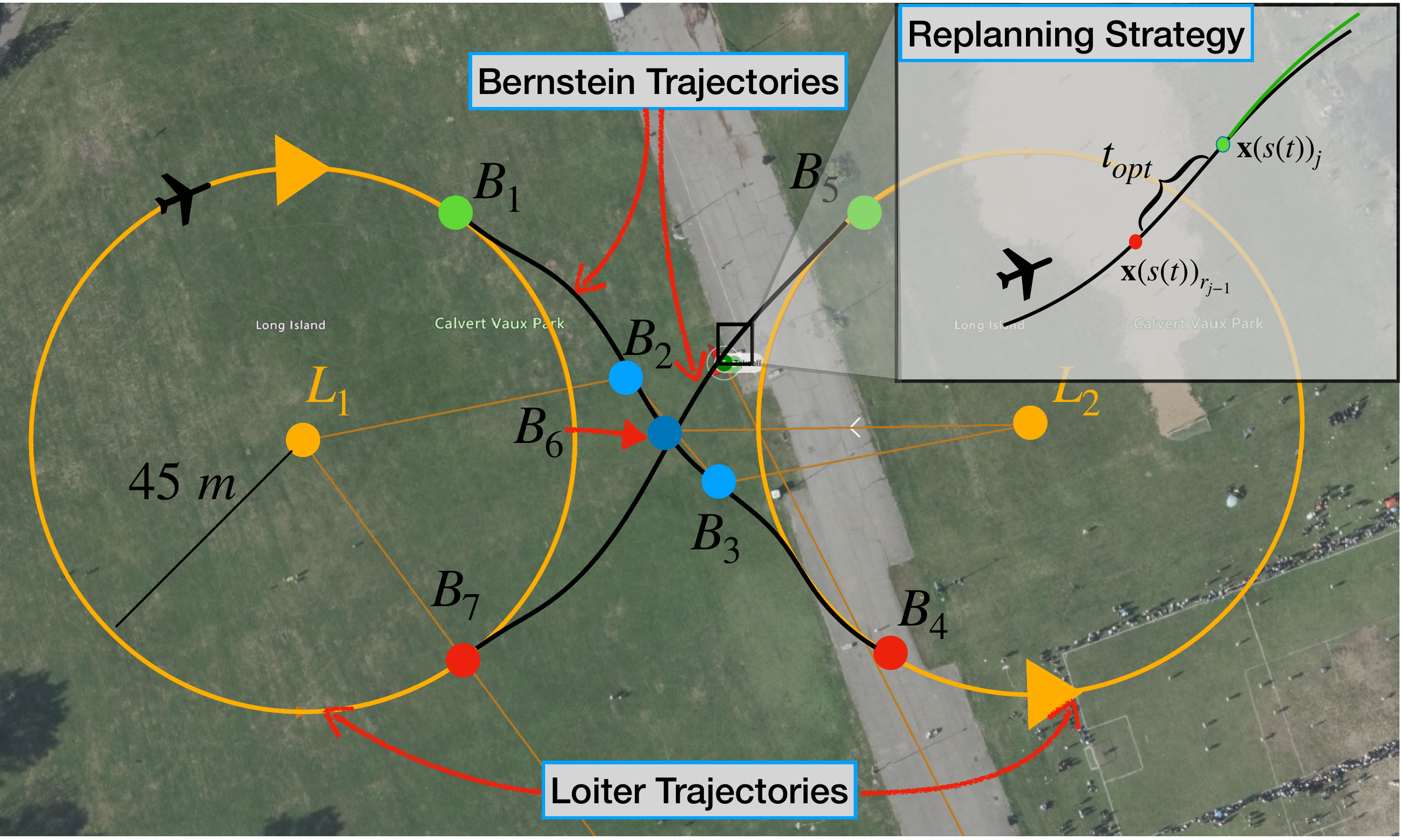}
  \caption{Planned Mission: Two loiter trajectories (yellow) and centered in $L_1$ and $L_2$ are linked together by two Bernstein trajectories passing through the set of waypoints $S_{\mathcal{B}_1} = [B_1, \dots, B_4]$ and $S_{\mathcal{B}_2} =[B_5, \dots, B_7]$ with a visualization of the replanning strategy (top right). 
  \label{fig:planned_mission}}
  \vspace{-10pt}
\end{figure}

\begin{enumerate}
    \renewcommand{\labelenumi}{\roman{enumi}.}
    \item \textit{Endpoint constraint:}
    Considering a starting time $t_0$ and an ending time $t_{f}$, we constrain $\mathbf{x}_{r}$ at the reference waypoints position $\mathbf{x}_{r}$, velocity $\dot{\mathbf{x}}_{r}$, and acceleration $\ddot{\mathbf{x}}_{r}$
     \begin{equation}\begin{aligned}
        C_{n,0}^{(k)}(t_0) = \mathbf{x}^{(k)}(t_0), \qquad C_{n,M}^{(k)}(t_f) = \mathbf{x}^{(k)}(t_f) 
    \end{aligned}.\end{equation}

    \item \textit{Continuity Constraints:}
    %Given a set $S_{\mathcal{B}}$ of $m_{i+1}$ waypoints, defined by a starting and ending time $t_{{m}_0}$ $t_{{m}_f}$,
    The goal is to ensure the continuity in position and higher derivatives of the trajectory $\mathbf{x}_r(t)$ at the junction of the $M$ sub trajectories as

    % \begin{equation}\begin{aligned}
    % \mathbf{C_m}(t_f) = \mathbf{C_{m+1}}(0) \\
    % \left\|  \mathbf{\dot{C}_m}(t_f) \right\| = \left\| \mathbf{\dot{C}_{m+1}}(0) \right\| \\
    % \left\|  \mathbf{\ddot{C}_m}(t_f) \right\| = \left\| \mathbf{\ddot{C}_{m+1}}(0) \right\|
    % \end{aligned}\end{equation}

    \begin{equation}\begin{aligned}
        C_{n,m}(t_{f}) = C_{n,m+1}(t_{{0}}). \\
    \end{aligned}\end{equation}

    \item \textit{Dynamic feasibility Constraints:}
    Given the FW dynamics, the curvature $\kappa = f(\dot{\mathbf{x}}_{r_x}, \dot{\mathbf{x}}_{r_y}, \ddot{\mathbf{x}}_{r_x}, \ddot{\mathbf{x}}_{r_y})$ evaluated from $t_0$ to $t_f$ of a given trajectory, needs to be constrained for its entire duration within the range  $\kappa_{min} \leq \kappa \leq \kappa_{max}$ to be considered feasible in order to avoid exceeding the maximum roll angle of the aircraft. Due to the non-linear nature of the curvature function $\kappa$, we apply a Taylor expansion around the equilibrium point to linearize the constraint, allowing us to maintain the original convex optimization problem formulation. The constraint  $k$ on the lineared curve is
    
    %The curvature $\kappa$ is a nonlinear function of $v_x, v_y, a_x$ and $a_y$ as expressed in the following equation:

    %\begin{equation}\begin{aligned}
   % k = f(v_x, v_y, a_x, a_y) = \frac{v_xa_y - a_xv_y}{(v_x^2 + v_y^2)^{3/2}}
    %\label{eq:curvature_k}
    %\end{aligned}\end{equation}

    \begin{equation}
    \begin{split}
        % &\phantom{=} k_{min} \leq k \leq k_{max} \\
        & \kappa_{min} \leq 
f(\dot{\mathbf{x}}_{r_x}, \dot{\mathbf{x}}_{r_y}, \ddot{\mathbf{x}}_{r_x}, \ddot{\mathbf{x}}_{r_y}) + \\
        &\begin{bmatrix} 
            \frac{\partial f}{\partial  \dot{\mathbf{x}}_{r_x}} &  \frac{\partial f}{\partial  \dot{\mathbf{x}}_{r_y}} & 
            \frac{\partial f}{\partial  \ddot{\mathbf{x}}_{r_x}} & \frac{\partial f}{\partial  \ddot{\mathbf{x}}_{r_y}}
        \end{bmatrix}
        \begin{bmatrix} 
            \dot{\mathbf{x}}_{r_x} - \dot{\mathbf{x}}_{r_x}(t_{rp}) \\\dot{\mathbf{x}}_{r_y} - \dot{\mathbf{x}}_{r_y}(t_{rp}) \\ \ddot{\mathbf{x}}_{r_x} - \ddot{\mathbf{x}}_{r_x}(t_{rp})  \\ \ddot{\mathbf{x}}_{r_y} - \ddot{\mathbf{x}}_{r_y}(t_{rp}) 
        \end{bmatrix} 
        \leq \kappa_{max}.
    \end{split}
    \label{eq:curvature}
    \end{equation}
    where $t_{rp} \in [t_0, t_f]$ represents the time instant where the linearization is applied.
    In particular, for a continuous linearization of the entire trajectory around a local point, a replanning strategy visible in Fig. \ref{fig:planned_mission} (top right) is applied at constant intervals. To avoid discontinuities between the current trajectory $\mathbf{x}_{r, j-1}$ and new replanned trajectory $\mathbf{x}_{r, j}$, we account for the optimization time $t_{opt}$ such that $\mathbf{x}_{r, j}(t_0) = \mathbf{x}_{r, j-1}(t + t_{opt})$.
    
    %The Taylor series approximation of the nonlinear constraint $k$ is accurate near the equilibrium point but insufficient for the entire trajectory. To address this problem, we implemented periodic replanning to maintain the $k$-constaint throughout the trajectory. As shown in Figure \ref{fig:planned_mission} (upper right), the replanning strategy is applied at constant intervals. In order  This transition point ensures continuity, allowing the aircraft to seamlessly switch to the new replanned trajectory when it reaches the $\rho(s(t))_j$. The real-world implementation and effectiveness of this strategy are discussed in the experimental results section. 

\end{enumerate}

\section{Results}
\label{sec:Experimental_Results}

In this section, we outline the mission planning for the fixed-wing using the combination of Bernstein polynomials and circular loiter trajectories. We conduct multiple experiments in a simulation environment, as shown in Fig.~\ref{fig:software_architecture} and in real-world settings in our outdoor flying arena located at Calvert Park in New York City. The design of the platform is inspired by the system proposed in \cite{wuest2022accurate}. The FW aircraft used in our experiments, shown in Fig. \ref{fig:fig1}, is a Hobby King Bixler 3 model, which is equipped with a Holybro\textsuperscript{{\textregistered}} PX4\footnote{\url{https://px4.io/}} Pixhawk autopilot for low-level attitude controller. Onboard computation is handled by an NVIDIA\textsuperscript{{\textregistered}} Jetson Xavier Orin board, running Ubuntu 20.04 and the ROS\footnote{\url{www.ros.org}} framework for intra-processes communications. For localization, we use a Drotek\textsuperscript{{\textregistered}} F9P GNSS receiver module, integrated with the PX4 EKF2-based state estimator.
We modeled the aerodynamic parameters described in Section \ref{sec:Aerodynamics_and_prop} following the approach outlined in \cite{Beard_book}. The planner and trajectory manager, as illustrated in Fig. \ref{fig:software_architecture}, operate at $100~\si{Hz}$ to ensure smooth and continuous control. Mission data is transmitted to the trajectory manager via the QGroundControl\textsuperscript{{\textregistered}} interface, running on a separate ground station. The interface also provides a real-time visualization of the vehicle’s status. %through telemetry data.  

\begin{figure}[!t]
  \centering
  \includegraphics[width=\columnwidth]{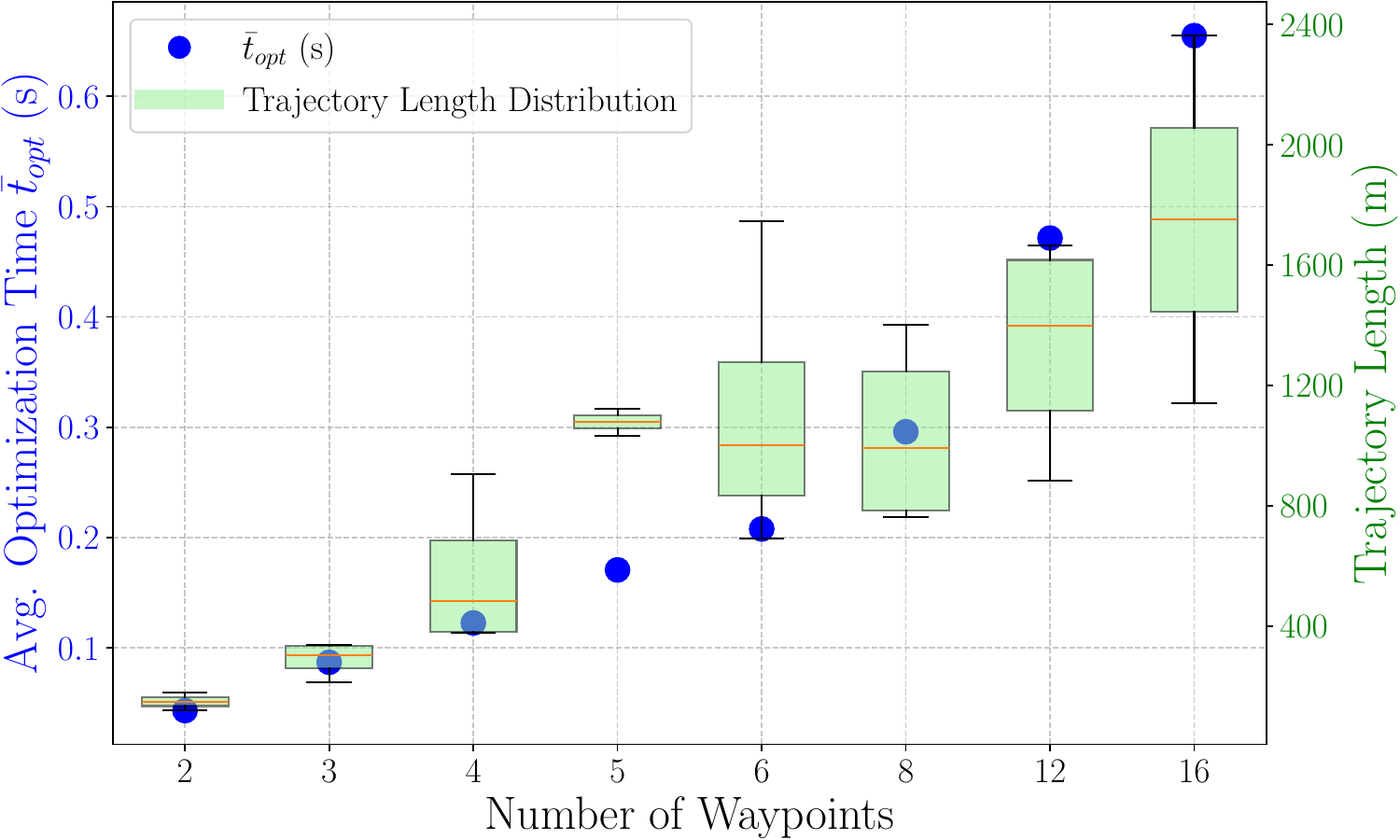}
  \caption{The average optimization computation time $\bar{t}_{opt}$ increases linearly with the number of waypoints remaining unaffected by the trajectory length.
  \label{fig:optimization_time}}
  \vspace{-20pt}
\end{figure}

\begin{figure*}[!t]
  \centering
  \includegraphics[width=1.0\textwidth]{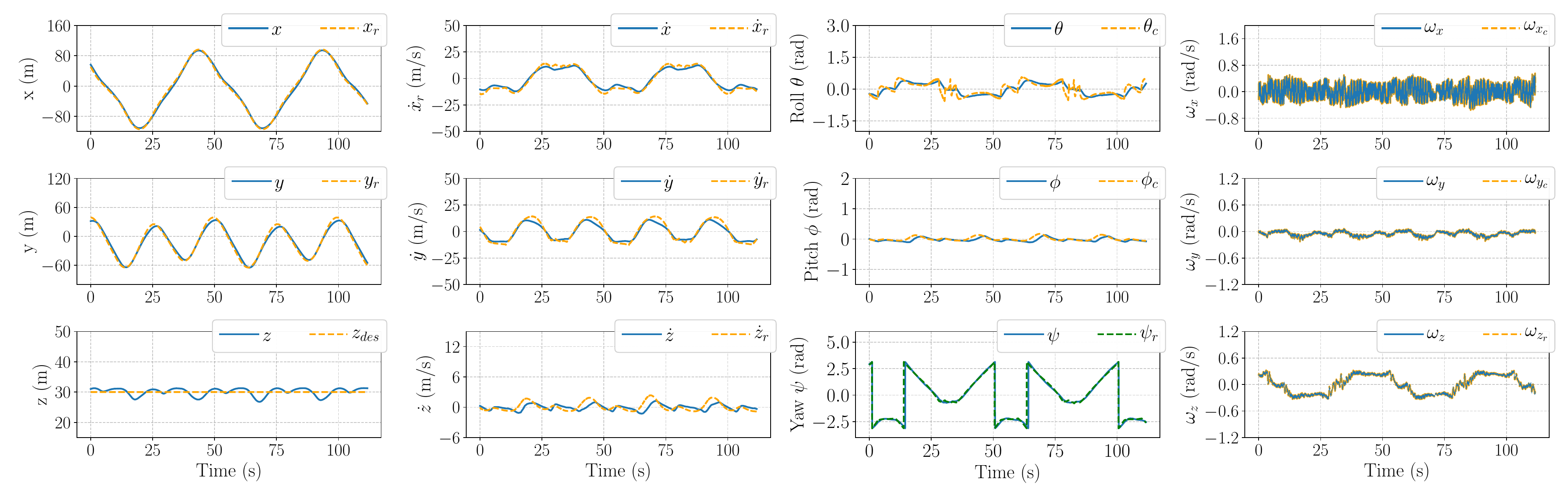}
    \vspace{-20pt}
  \caption{Simulation results obtained using the SITL PX4 simulator with Gazebo as the physics engine with a planned mission similar to Fig.~\ref{fig:planned_mission}. In $2~\si{m/s}$ NE wind conditions. The results show good position and attitude tracking performance. The yaw $\psi$ (green) is not a control variable. 
  \label{fig:experiment_result}}
  \vspace{-10pt}
\end{figure*}

\begin{figure*}[!t]
  \centering
  \includegraphics[width=1.0\textwidth]{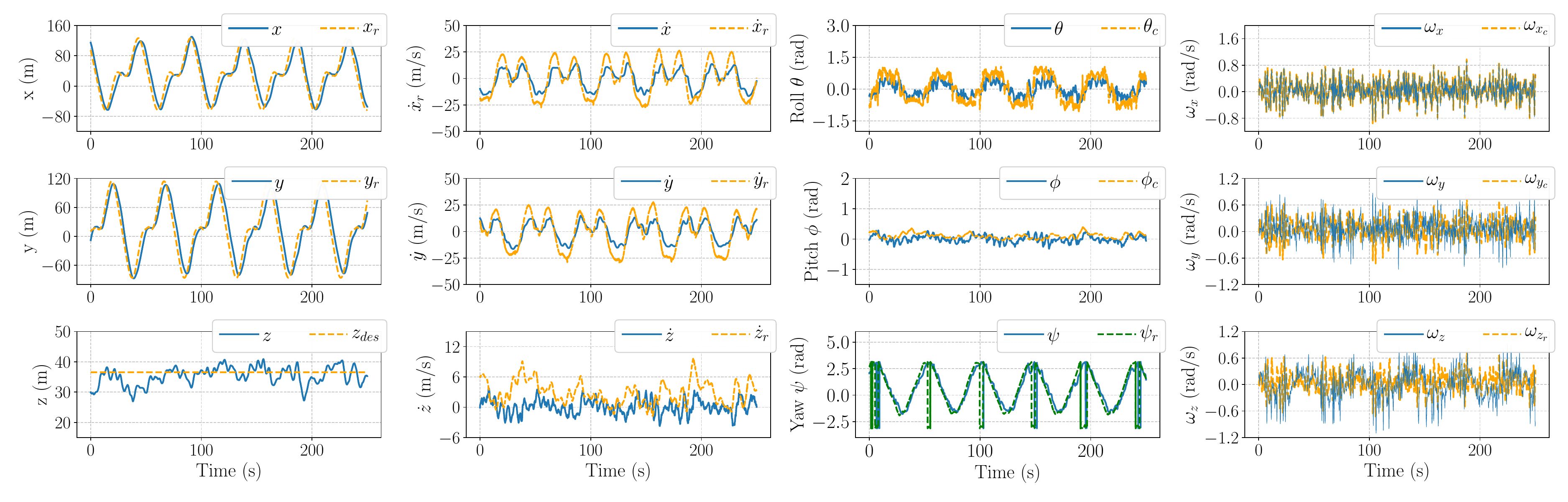}
    \vspace{-20pt}
  \caption{Results obtained in real experimentation with our platform, leveraging the differential flatness equations and forwarding desired attitude commands to an onboard PX4 controller, in the mission proposed in Fig. \ref{fig:planned_mission}. Despite a recorded wind with $3.4~\si{m/s}$, the results show a very close sim-to-real gap with good tracking capability of the reference trajectory.
  \label{fig:real_experiment_result}}
  \vspace{-20pt}
\end{figure*}

%To put iterations instead of traj
\begin{figure}[t]
  \centering
  \includegraphics[width=1.0\columnwidth]{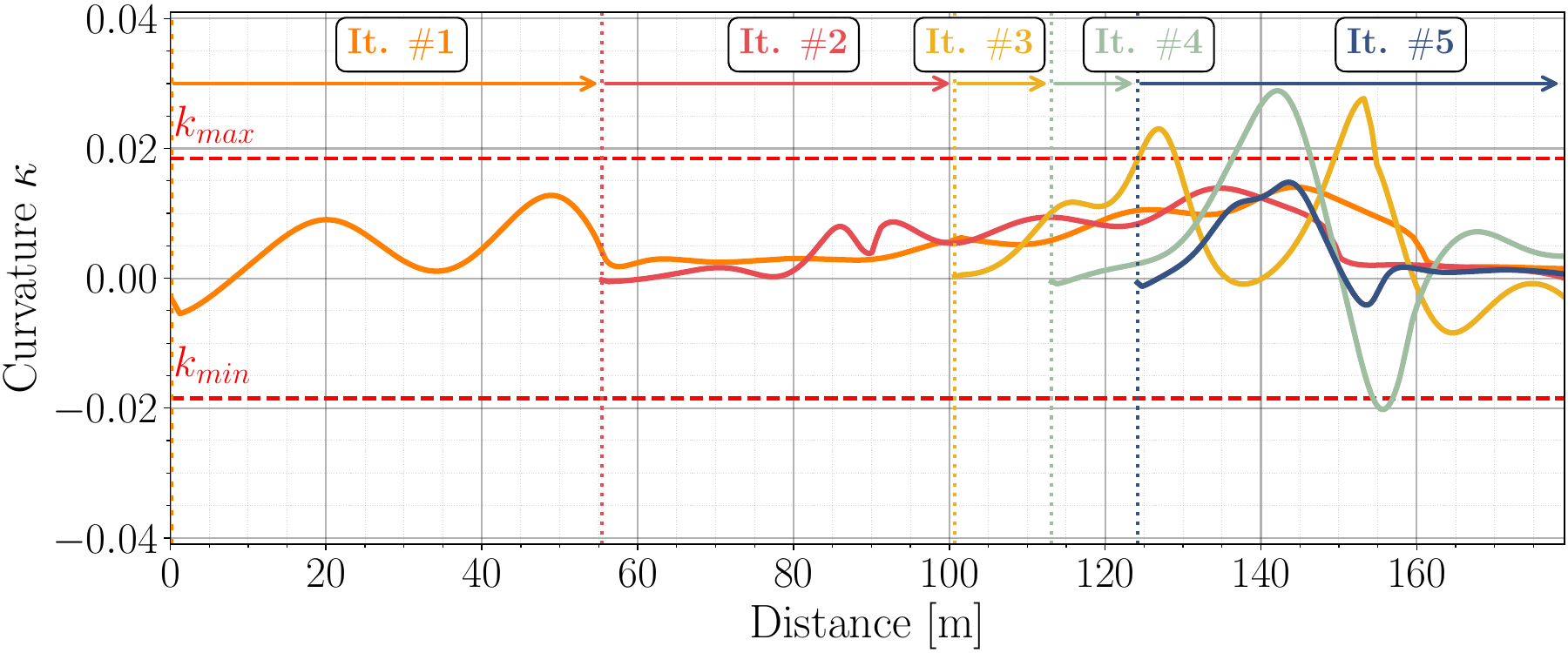}
  \caption{Bernstein Trajectory replanning over multiple iterations for curvature constraint feasibility. 
  \label{fig:curvature_k_result}}
  \vspace{-20pt}
\end{figure}

\subsection{Mission Planning}
For both simulation and real-world experiments, we plan the mission using the QGroundControl (QGC) interface as visible in Fig. \ref{fig:planned_mission}. The waypoints are uploaded to the onboard computer, referencing map coordinates, with each waypoint serving a specific purpose. We designed an experiment that combined multiple consecutive trajectories. The loiter waypoints, denoted by $L_1$ and $L_2$, generate circular loiter trajectories depicted in yellow. Two additional intermediate Bernstein polynomial trajectories are generated through Bernstein waypoints ($B_1, B_2, B_3, B_4$) and ($B_5, B_6, B_7$), visible in black. 
The trajectory is optimized online using the C++ OSQP library~\cite{osqp}, which computes kilometers long trajectories in less than a second, as visible in Fig.~\ref{fig:optimization_time}, where an average optimization time $\bar{t}_{opt}$ has been computed between $4$ consecutive optimization run with an increasing waypoints number. 
In the setup proposed in Fig. \ref{fig:planned_mission}, we measure $\bar{t}_{opt} = 0.083~\si{ms}$. To impose the non-linear constraint $\kappa$, we run a continuous trajectory replanning at $10~\si{Hz}$. 

%In Section \ref{section:sim_results}, we present the simulation results. Then, in Section \ref{section:rw_results}, we provide the outcomes from performing the same mission in a real-world scenario.

\subsection{Simulation Results}
\label{section:sim_results}
We validate our proposed solution in multiple simulation experiments. To decrease the sim-to-real gap, we leverage the PX4 SITL simulator, which provides the possibility of accurate simulated fixed-wing dynamics and the advantage of planning the mission directly in the location where the real tests are going to be performed. As visible in Fig. \ref{fig:planned_mission}, the mission is composed of two loiter trajectories with a radius $r = 45~\si{m}$ each and connected at the tangential points by two Bernstein trajectories. The entire length of the trajectory is $1143.24~\si{m}$, and a wind of $2~\si{m/s}$ has been simulated in the NE direction. 
The trajectory tracking results are shown in Fig.~\ref{fig:experiment_result}. We define with $RMSE_{pos}$ and $RMSE_{vel}$  the combined Root Mean Square Error across the three Cartesian directions for positions and velocities, respectively, which is $RMSE_{pos} = 6.031~\si{\meter}$ and $RMSE_{vel} = 3.316~\si{m/s}$. 
The maximum and minimum values of roll recorded during the overall trajectory length are $\phi_{max} = 0.572~\si{rad}$ and $\phi_{min} = 0.561~\si{rad}$ respectively.
An example of the proposed replanning technique along the trajectory connecting waypoint $B_5 - B_7$  is visualized in Fig. \ref{fig:curvature_k_result}. As visible, after a distance of $d = 120~\si{m}$ the curvature of the trajectory computed at iteration $i= 1$ exceeds the desired value of $\kappa=\pm 0.02$, but the continuous replanning strategy ensures that the trajectory computed at $i = 5$ is shaped respecting locally the curvature constraint presented in eq.~\eqref{eq:curvature}.

\subsection{Real World Experiments}
\label{section:rw_results}
Next, to validate the performances of our control and planning solutions, we conduct several experiments in real-world scenarios, leveraging our in-house developed platform visible in Fig. \ref{fig:fig1} and introduced in Section \ref{sec:Experimental_Results}.

The results are illustrated in Fig. \ref{fig:real_experiment_result}, showing the aircraft's behavior during a 5-minute experiment. The total distance traveled is 3256.81~\si{m}, with a desired airspeed of $Va = 14~\si{m/s}$. The mission setup is similar to that depicted in Fig. \ref{fig:planned_mission} and discussed in the previous section. During the experiment, wind conditions are at an intensity of 3.4~\si{m/s} from a southeast direction. 
As noticeable, our approach provides good planning and tracking results showing similar performance with respect to simulation results with a global tracking error in positions and velocities of $RMSE_{pos} = 13.441~\si{m}$ and $RMSE_{vel} = 10.895~\si{m/s}$ respectively.
Moreover, during the test, the Bernstein trajectories are continuously generated by the optimizer, requiring an average optimization time $\bar{t}_{opt} =  0.0623~\si{s}$, proving the reliability and the applicability of our method also in continuous real-time flight conditions. 
During the experiment, the aircraft achieves a maximum and minimum value of a roll of $\phi_{max} = 0.87~\si{rad}$ and $\phi_{min} = -0.649~\si{rad}$ respectively. Our successful experimental results prove the applicability and reliability of the proposed approach for controlling FW-UAVs in real scenarios.

\section{Conclusion}
\label{sec:Conclusion}
In this paper, we proposed a complete real-time planning and control approach for continuous, reliable, and fast online generation of dynamically feasible Bernstein trajectories and control for FW aircrafts. The generated trajectories span kilometers, navigating through multiple waypoints. By leveraging differential flatness equations for coordinated flight, we ensure precise trajectory tracking. Our approach guarantees smooth transitions from simulation to real-world applications, enabling timely field deployment. The system also features a user-friendly mission planning interface. Continuous replanning  maintains the rajectory curvature 
$\kappa$ within limits, preventing abrupt roll changes.

Future works will include the ability to add  a higher-level kinodynamic path planner to optimize waypoint spatial allocation and improve replanning success, and enhancing the trajectory-tracking algorithm by refining the aerodynamic coefficient estimation.

%\addtolength{\textheight}{-9cm}   % This command serves to balance the column lengths

%%%%%%%%%%%%%%%%%%%%%%%%%%%%%%%%%%%%%%%%%%%%%%%%%%%%%%%%%%%%%%%%%%%%%%%%%%%%%%%%

\bibliographystyle{IEEEtran}
\bibliography{reference}

\end{document}